\documentclass[letterpaper, 10 pt, conference]{ieeeconf}
\IEEEoverridecommandlockouts
\usepackage{cite}
\usepackage{amsmath,amssymb,amsfonts}
\usepackage{algorithmic}
\usepackage{graphicx}
\usepackage{blindtext}
\usepackage{textcomp}
\usepackage{xcolor}
\usepackage{siunitx}
\usepackage{wrapfig}
\usepackage{textcomp}
\usepackage{cite,url,verbatim,color,comment}
\usepackage{balance}
\usepackage{xcolor}
\usepackage[colorlinks=true]{hyperref}

    \hypersetup{backref=true,       
                    pagebackref=true,               
                    hyperindex=true,                
                    colorlinks=true,                
                    breaklinks=true,                
                    urlcolor= black,                
                    linkcolor= blue,                
                    bookmarks=true,                 
                    bookmarksopen=false,
                    filecolor=black,
                    citecolor=blue,
                    linkbordercolor=blue
}

\title{\LARGE \bf A Framework for Controlling Multi-Robot Systems Using\\ Bayesian Optimization and Linear Combination of Vectors\\
\thanks{This work was supported in part by the National Science Foundation Research Experience for Undergraduates Program, Award \#1851815, and
West Virginia University Statler College.}
}

\author{Stephen Jacobs \and R. Michael Butts \and Yu Gu \and Ali Baheri \and Guilherme A. S. Pereira\thanks{The authors are with the Department of Mechanical and Aerospace of the Benjamin M. Statler College of
Engineering and Mineral Resources at West Virginia University, Morgantown, WV, USA. Emails: sej0015@mix.wvu.edu, rmb0034@mix.wvu.edu, yu.gu@mail.wvu.edu, ali.baheri@mail.wvu.edu, and guilherme.pereira@mail.wvu.edu.}}

\begin{document}
\maketitle

\begin{abstract}
We propose a general framework for creating parameterized control schemes for decentralized multi-robot systems. A variety of tasks can be seen in the decentralized multi-robot literature, each with many possible control schemes. For several of them, the agents choose control velocities using algorithms that extract information from the environment and combine that information in meaningful ways. From this basic formation, a framework is proposed that classifies each robots' measurement information as sets of relevant scalars and vectors and creates a linear combination of the measured vector sets. Along with an optimizable parameter set, the scalar measurements are used to generate the coefficients for the linear combination. With this framework and Bayesian optimization, we can create effective control systems for several multi-robot tasks, including cohesion and segregation, pattern formation, and searching/foraging.

\end{abstract}

\section{Introduction}

Multi-robot systems provide several unique challenges when compared to their single-agent counterparts, but also create new possibilities for interesting applications and behaviors.  Because of this, research within the field of decentralized multi-robotic systems has increased, resulting in several specialized controllers being proposed.
Often, these controllers are not able to be generalized to produce behaviors beyond the robot's primary task. This limitation can be seen, for example, when we try to combine control schemes for collision avoidance with other controllers to create a secondary desired behavior~\cite{taylor2020emergent}. Because of this limitation on current control schemes, it would be beneficial to develop a general framework for creating parameterized control schemes for multi-robot systems that use measured scalars and vectors to produce desired behaviors. 

By introducing the control framework proposed in this paper we hope to simplify the construction of decentralized multi-robot controllers. With an understanding of the desirable and undesirable outcomes of a multi-robot system and an intuition about what sensor information is important to the robots for decision making, a control structure can be created using the proposed framework. The parameters of the control structure can then be optimized for different cost functions to vary the agents' behavior. The proposed framework also provides a starting point for future research regarding the importance of robots' measurements to their behaviors. Therefore, the main contribution of this paper is a general, optimizable framework that can encapsulate several multi-robot, velocity-based controllers. The paper shows several examples of applications where the proposed framework can be used. We start with a literature review that surveys and describes these applications.

\begin{figure}[t]
   \centerline{\includegraphics[width=7.0cm,trim={0 0 0 0},clip]{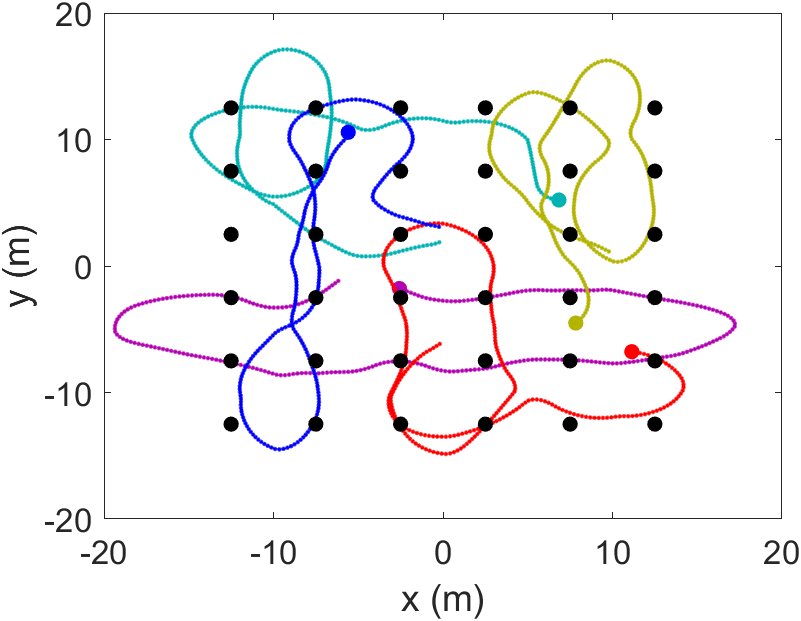}}
    \caption{Trajectories of a five-agent decentralized searching behavior generated using the proposed control framework.}
    \label{fig:hand_search}
\end{figure} 

\section{Literature Review}
 The framework proposed in this paper is designed specifically to support decentralized robotic systems. Many of the advanced multi-robot systems currently operating in industry run in a centralized manner, possibly with a partially distributed control scheme. Within centralized systems, it is often possible to use long-term planning to find optimal solutions to a variety of problems. With decentralized systems, however, limitations on sensing range and communication make finding optimal solutions challenging. This leads to the development of control schemes based on human intuition that are sensitive to parameter tuning. To fully understand this problem, it is important to analyze different currently used control schemes. In this section, proposed methods for some decentralized multi-robot tasks will be discussed and their solutions compared. 

\subsection{Flocking}
Flocking has long been a task proposed for multi-robot systems. Inspired by biological systems, flocking behavior involves agents grouping together, attempting to maintain a defined inter-agent separation, and aligning their headings with one another. One of the most famous flocking algorithms is the Boids algorithm proposed by Reynolds in 1987~\cite{reynolds1987flocks} and further analyzed in~\cite{tanner2007flocking}. 
The beauty of the Boids algorithm lies in its simplicity: with only three control rules, the agents can form convincing biological structures that change and move dynamically, and through tuning, the behavior can be changed continuously, thereby expressing a large continuum of forms. In the implementation of the Boids algorithm, the agents are under the influence of three forces/torques, computed locally: a force that points towards the “center of mass” of neighboring agents, a repelling force from nearby agents, and a torque that aligns its velocity vector with the velocities of neighboring agents. Each of these artificial forces are developed by individual agents' measurement of neighboring agents’ positions and velocities. The strengths of the forces are also determined by relations of those measurements, such as the relative distance to each neighboring agent and some constant parameters.

\subsection{Cohesion and Segregation}
The cohesion and segregation task is in some ways an extension of flocking. The two parts of the task are cohesion: similar agents (e.g. same “group”, color, shape, etc) come together, and segregation: dissimilar agents (e.g. different “group”, etc.) move apart. The end goal is for robots in each group to be together and for robots in two different groups to be separated. With centralized control, this task can be done very efficiently~\cite{santos2020spatial}, but for decentralized agents with only local sensing, the task becomes more challenging. Rezec et al~\cite{rezeck2021flocking} propose an interesting approach that uses potential fields and the construction of a Gibbs Random Field to effectively achieve cohesion and segregation with a large number of agents using only local sensing. In the method, a potential field is defined that has two main parts: attraction and repulsion from other agents, and kinetic energy matching. The agents are essentially attracted to neighboring agents of the same group, repelled from neighbors of different groups, and encouraged to take velocities similar to the velocities of neighbor agents in their group. This is done using a Gibbs Random Field \cite{sherman1973markov} and the Metropolis-Hastings algorithm \cite{robert1999metropolis} to select new velocities.

\subsection{Formation Control}
It is easy to imagine scenarios where formation control would be desirable in multi-robot systems. By assembling into different structures, agents can interact with their surroundings in more interesting ways. One way to design a control algorithm for decentralized agents is using potential or  navigation functions~\cite{rimon1990navigation}. In \cite{de2006formation}, for example, the agents are steered using a navigation function to create formations based on pre-prescribed distances between each agent. The agents measure their relative positions and distances to one another and to the obstacles and use that information to construct individual navigation functions on which to traverse. The construction of the navigation function requires defining several component functions such as the “goal function” and “obstacle function” with the correct shapes to encourage efficient formation behavior, but which are inflexible to changes in their task.

\subsection{Other applications}
Methods for pattern formation~\cite{flocchini2008arbitrary}, foraging~\cite{zedadra2017multi}, and  collision avoidance~\cite{van2011reciprocal} have also been developed for multi-robot systems. In each of these methods, the agents must obtain sensory information from their surroundings and use that information to plan an input control velocity. In each method, however, the sensory information is combined in different ways, so there is little crossover between one method and another. In the generic framework described in this article, we propose a method capable of creating complex multi-robot behaviors similar to those described above. While the framework may not be able to control agents for each task as efficiently as a specifically designed control method, its flexibility and ability to be optimized using off-the-shelf methods means that it can create a huge variety of cooperative tasks easily given the appropriate cost-function and a little creativity. Before we describe the proposed framework in Section~\ref{sect:framework}, the next section presents our problem definition.

\section{Problem Definition}
    We consider a multi-robot system with $N\geq 2$ holonomic mobile robots,  Each robot makes velocity control decisions independent of each other robot. Each robot's decisions are determined by their current observed state $S$, which is defined by what they perceive from the environment via local sensing or local communication. This state is saved locally as a set of scalars and vectors that may include values such as density of local agents, distance to the closest neighbor, vectors pointing to the centroid of the group or to a goal position, or any other meaningful quantity. Our problem is to combine such local information to generate a control velocity vector for each agent in such a way that the group of robots performs a desired task. The method should be generalizable to many tasks and optimizable, using a small number of parameters to define the control scheme. Tasks can be specified by a global cost function that must be minimized by the overall group of robots. The cost function can be seen as a function of user-defined parameters $P$ where each measured parameter $p\in P$ describes some aspect of the robot-environment interactions and is multiplied by some set multiplier $K$ as well as a value $a$, where $a = \pm 1 $ based on if the parameter is desirable ($-$) or undesirable ($+$). Then, these parameters are summed over a time step until time $T$. This cost function can be written as:      
    \begin{equation}
        \label{eqt:cstfunc}
        C(p) = \sum_{t=0}^{T}\Big( \sum_{i=1}^{P} a_i\cdot K_i\cdot p_{i_{\Delta_t}}\Big)\cdot \Delta t. 
    \end{equation} 
    To solve this problem, we propose a generic framework that is presented in the next section. 


\section{Proposed Framework}
\label{sect:framework}

To create the desired parameterized function of measured variables that will produce a generalizable and optimizable control function, two classifications of measurements are considered: scalars and vectors. This may exclude some forms of complex measurements, but most sensory data from mobile robots can be broken down into these categories or the data can be processed slightly to create meaningful quantities that fit into these categories. Assuming a meaningful set of scalars and vectors, it is reasonable to assume that for any given cost function, there is some function on the scalars that could create coefficients for a combination of the vectors to form a control vector function as shown in~ (\ref{eqt:controlvectorfunction}), which could produce a good solution. 

\begin{equation}
    \begin{matrix}
        F_{1}(S_{1},S_{2},\cdots, S_{n})\overrightarrow{V_{1}}\\ 
        +\\ 
        F_{2}(S_{1},S_{2},\cdots, S_{n})\overrightarrow{V_{2}}\\ 
        +\\ 
        \cdots\\ 
        +\\ 
        F_{m}(S_{1},S_{2},\cdots, S_{n})\overrightarrow{V_{m}}
    \end{matrix}
    =
    \overrightarrow{V_{c}}
    \label{eqt:controlvectorfunction}
\end{equation}

 While the optimal scalar-valued functions for any given cost function are very likely non-linear, we propose linearizing them to decrease the number of parameters and increase the generalizability.
The proposed linearized function shown in (\ref{eqt:linsyst}) consists of an $m\times n$ matrix of parameters multiplied by an $n\times 1$ vector of measured scalars. This results in an $m\times 1$ vector of coefficients that are used to combine the measured vectors using the dot product operator. The resulting vector is then taken as the control velocity for the agent.
\begin{equation} 
\label{eqt:linsyst}
    \begin{bmatrix}
        p_{11} & p_{12} & \ldots & p_{1n}\\ 
        p_{21} & p_{22} & \ldots & p_{2n}\\ 
        \vdots & \vdots & \ddots & \vdots\\ 
        p_{m1} & p_{m2} & \ldots & p_{mn}
    \end{bmatrix}
    \times 
    \begin{bmatrix}
        S_{1}\\ 
        S_{2}\\ 
        \vdots\\ 
        S_{n}
    \end{bmatrix}
    \cdot 
    \begin{bmatrix}
        \overrightarrow{V_{1}}\\ 
        \overrightarrow{V_{2}}\\ 
        \vdots\\ 
        \overrightarrow{V_{n}}
    \end{bmatrix}
    = 
    \overrightarrow{V_{c}}
\end{equation}

The exact set of measurement data (scalars and vectors) used in the previous equation is dependent on the desired task. Many multi-robot tasks could likely be performed using the same set of measurements, but some specific tasks may need more or fewer measurements as inputs. Some examples of vector measurements that could be used for multiple tasks are the unit vector from the central agent towards the nearest neighboring agent, the unit vector to some goal or goal object, the unit vector pointing towards the center of mass of the local group of agents, or the unit vector pointing in the direction of the average velocity of the neighboring agents. On the other hand, important scalar measurements may include the distance to the nearest neighbor, the distance to a goal position, the number of other agents in the local sensing range, or the average velocity magnitude of the neighboring agents.
\begin{figure}[ht]
   \includegraphics[width=0.49\linewidth]{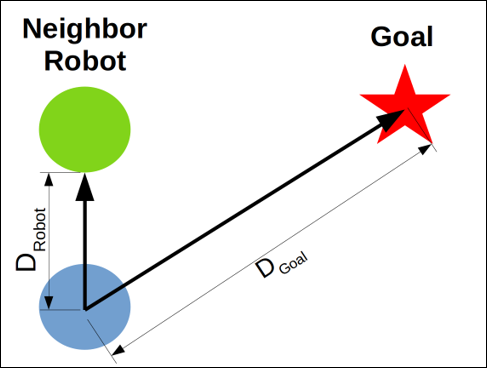}
   \hfill
   \includegraphics[width=0.49\linewidth]{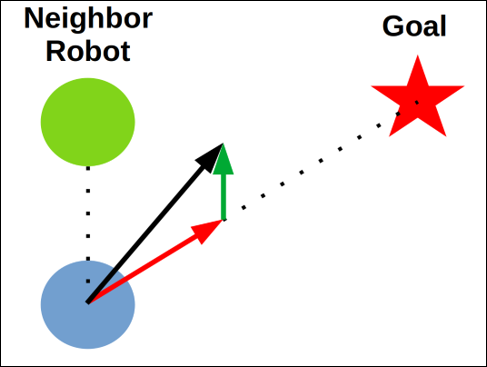}
    \caption{Example application of the proposed methodology. The robots use two scalars ($D_{goal}$ and $D_{robot}$) and two vectors (unitary vector to the goal, in red, and unitary vector to the neighbor, in green) to compose the final control vector using (\ref{eqt:linsyst}). The parameters in this equation will depend on the required robot behavior.}
    \label{fig:examples_scalars_vectors}
\end{figure} 

Fig.~\ref{fig:examples_scalars_vectors} shows an example of a multi-robot task and how the framework could be applied. In this example, the blue agent’s control structure consists of two scalars and two vectors. The scalars are the distance towards the nearest neighbor and the distance to the goal. The two vectors used in the control structure are the unit vector towards the nearest neighbor and the unit vector towards the goal. The control velocity for an agent is found as a linear combination of the measured vectors, using the scalars and parameters to find the appropriate lengths as:
\begin{equation}
\label{eqt:worked_out}
\begin{aligned}
    \begin{bmatrix}
        p_{11} & p_{12}\\ 
        p_{21} & p_{22}
    \end{bmatrix}
    \times
    \begin{bmatrix}
        S_{neig.}\\
        S_{goal}
    \end{bmatrix}
    \cdot
    \begin{bmatrix}
        \overrightarrow{V}_{neig.}\\
        \overrightarrow{V}_{goal}
    \end{bmatrix}
    =
    \overrightarrow{V}_{c}\\
    \begin{matrix}
        (p_{11}S_{neig.}+p_{12}S_{goal})\overrightarrow{V}_{neig.}\\
        +\\
        (p_{21}S_{neig.}+p_{22}S_{goal})\overrightarrow{V}_{goal}\\
    \end{matrix}
    =
    \overrightarrow{V_{c}}
\end{aligned}
\end{equation}

While this example is fairly simple and the four parameters could be easily tuned by hand to function in the desired final behavior, as the number of parameters increases, the optimal solution becomes difficult to find by manually tuning the parameters. By using Bayesian optimization~\cite{calandra2014bayesian,baheri2017real,berkenkamp2021bayesian,nogueira2016unscented}, a much greater number of parameters can be tuned effectively allowing the control structure to have an expanded number of included measurements, which leads to more complex behaviors and a better solution to the given cost function.

Since the relation between the measurements and the cost metric is not known \emph {a priori}, we cast the tuning problem as a black-box optimization problem and leverage Bayesian optimization for solving it. Bayesian optimization seeks to find the global optimum of a black-box function within only a few function evaluations. One popular approach is to model the unknown function as a Gaussian process~\cite{rasmussen2003gaussian}, where Bayesian optimization puts prior belief in the structure of that cost metric. In particular, at every step, the next evaluation point is selected to maximize some acquisition function that characterizes (i)~how much will be learned by evaluating a set of measurements 
and (ii)~what the likely performance level will be at that next set of measurements.

\section{Example Applications and Experiments}

\subsection{Flocking}

\begin{figure}[tbp]
   \centerline{\includegraphics[width=0.8\linewidth]{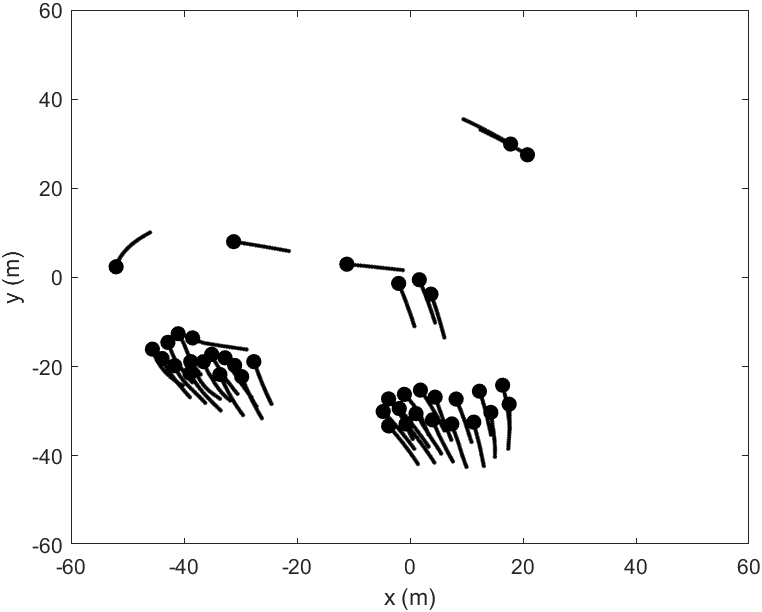}}
    \caption{Flocking behavior of forty agents controlled by the control framework shown in (\ref{eqt:flock_struct}) after random initialization.}
    \label{fig:flock1}
\end{figure}

A straightforward application of the proposed framework is flocking. For this application, the structure of the framework is composed of three scalar measurements -- the distance to the closest neighbor $S_{N}$, the distance to the origin of the map $S_{O}$, and a constant bias term --  and five measured vectors -- the unit vector pointing toward the centroid of neighboring agents (robots within a $4$\,m radius in our simulations)  $\overrightarrow{V}_{C}$, a unit vector pointing to the nearest neighbor agent $\overrightarrow{V}_{N}$, the unit vector pointing to the origin $\overrightarrow{V}_{O}$, the unit vector pointing in the direction of the average heading of nearby agents $\overrightarrow{V}_{H}$, and the current velocity vector $\overrightarrow{V}_{V}$. To improve the outcome of the framework, the scalar inputs were taken as basic functions of the scalar measurements as shown in (\ref{eqt:flock_struct}). We noticed that, for repulsion behaviors, the inverse of a distance measurement was often more effective than the measurement itself. In this case, the inverse of $S_N$ cubed was used as it showed interesting behavior, although similar behavior was also possible with the inverse only. Also, in these simulations, it was necessary to provide the robots an attractive force to the map's center so that they wouldn't leave the map area. To limit this effect to the edges of the map, the distance to the map origin, $S_{O}$, was divided by fifty (nearly the map size) and raised to the sixth so that the strength of the force would be negligible near the center of the map and increase dramatically towards the edges. Because there was not a cost function being optimized for flocking, the parameters were tuned by hand and a simple solution that produced satisfactory results was found as:
\begin{equation} 
\begin{bmatrix} 
        -50 & 0 & 0\\ 
        0 & 5 & 0\\
        0 & 0 & 0.5\\
        0 & 0 & 25\\
        0 & 0 & 10
    \end{bmatrix}
    \times 
    \begin{bmatrix}
        \frac{1}{(S_{N})^3}\\ 
        (\frac{S_{O}}{50})^6\\  
        1
    \end{bmatrix}
    \cdot 
    \begin{bmatrix} 
        \overrightarrow{V}_{N}\\
        \overrightarrow{V}_{O}\\
        \overrightarrow{V}_{C}\\
        \overrightarrow{V}_{H}\\
        \overrightarrow{V}_{V}
    \end{bmatrix}
    = 
    \overrightarrow{V_{c}}
    \label{eqt:flock_struct}
\end{equation}
An example of the behavior obtained when each robot follows the velocity computed by (\ref{eqt:flock_struct}) is shown in Fig.~\ref{fig:flock1}. The dynamic behavior of the robotic team can be observed in the accompanying video.

\vspace{-.5mm}
\subsection{Cohesion and Segregation}

\begin{figure}[t]
\centering
  \includegraphics[width=0.48\linewidth]{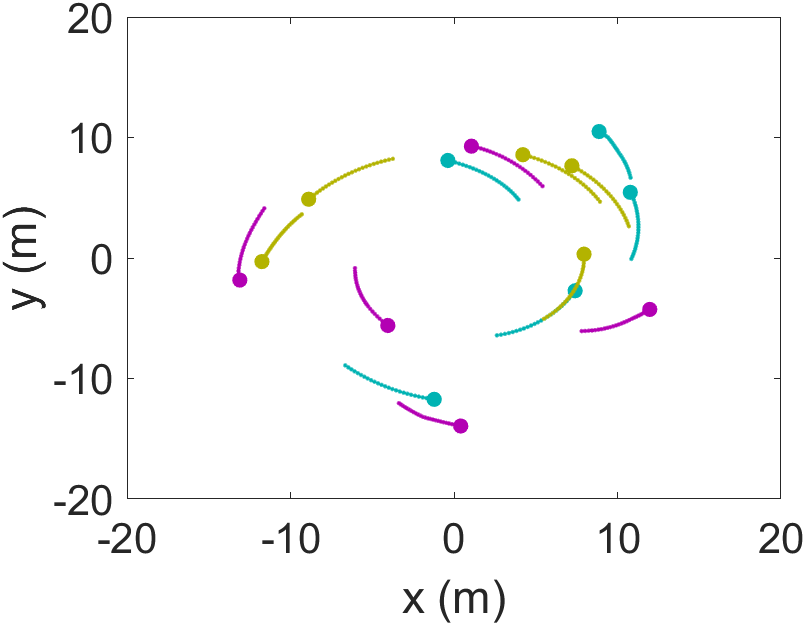}
  \includegraphics[width=0.48\linewidth]{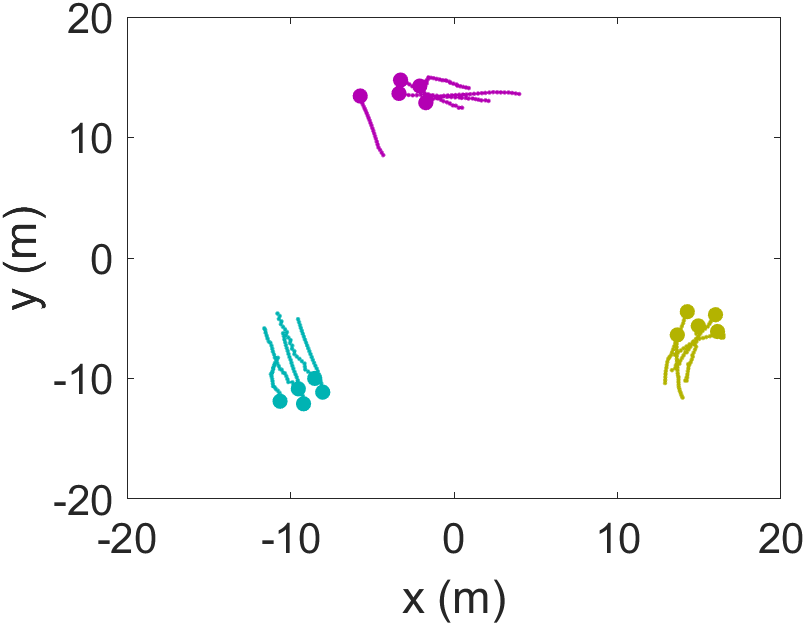}
    \caption{Two snapshots of cohesion and segregation behavior are exhibited among three groups of five agents each.}
    \label{fig:cohseg}
\end{figure}

Another instance of the framework was developed to show it can be used for cohesion and segregation. Although there are examples of cost functions for the cohesion and segregation task in the literature~\cite{rezeck2021flocking}, the measured scalars, vectors, and parameters in this section were also chosen using intuition. In order to produce the desired behavior, six scalars and six vectors were used in the control structure. To generate the measurements, two centroids are calculated from each agent’s measurement data: the centroid of neighboring agents ($25$\,m radius) in the same group and the centroid of neighboring agents in a different group. The unit vectors to each of these centroids ($\overrightarrow{V}_{1}$ and $\overrightarrow{V}_{3}$), as well as orthogonal unit vectors to these vectors ($\overrightarrow{V}_{2}$ and $\overrightarrow{V}_{4}$), are used as vector inputs to the structure along with a unit vector pointing to the nearest neighbor ($\overrightarrow{V}_{5}$) and a unit vector pointing towards the map’s origin ($\overrightarrow{V}_{6}$). For scalar inputs, the number of neighbors in the same group $S_{N_1}$, the number of agents in a different group $S_{N_2}$, the distance to the closest neighbor $S_{N}$, the distance to the centroid of agents from different groups $S_{diff}$, the distance to the map origin $S_{O}$, and a constant bias term were all included. As a more complex task than the flocking task, the hand-tuned parameter set also included more non-zero terms to produce the desired behavior as shown in~(\ref{eqt:coh_struct}). Similar to the flocking structure, powers of the inverses of two scalars were used instead of the scalars themselves in order to improve the repulsion behavior. Some results generated by this control structure are shown in Fig.~\ref{fig:cohseg} and the accompanying video. 
\begin{equation} 
\small
\setlength{\arraycolsep}{2pt}
    \begin{bmatrix}
        -0.3 & -0.3 & 0 & -3 & 0 & 5.4\\ 
        0.3 & 0 & 0 & 0 & 0 & -2.4\\ 
        0 & -0.3 & -3 & -3 & 0 & 0.9\\ 
        0 & 0 & -9 & 0 & 0 & 1.2\\ 
        0 & 0 & 0 & 0 & 0.03 & 0.03
    \end{bmatrix}
    \times 
    \renewcommand\arraystretch{1.2}
    \begin{bmatrix}
        S_{N_{1}}\\ 
        S_{N_{2}}\\
        \frac{1}{(S_{N})^2}\\
        \frac{1}{(S_{diff})^3}\\
        S_{O}\\
        1
    \end{bmatrix}
    \cdot 
    \renewcommand\arraystretch{1}
    \begin{bmatrix} 
        \overrightarrow{V}_{1}\\
        \overrightarrow{V}_{2}\\
        \overrightarrow{V}_{3}\\
        \overrightarrow{V}_{4}\\
        \overrightarrow{V}_{5}\\
        \overrightarrow{V}_{6}
    \end{bmatrix}
    = 
    \overrightarrow{V_{c}}
    \label{eqt:coh_struct}
\end{equation}

\vspace{-1mm}
\subsection{Pattern Formation}

\begin{figure}[htbp]
   \centerline{\includegraphics[width=0.75\linewidth]{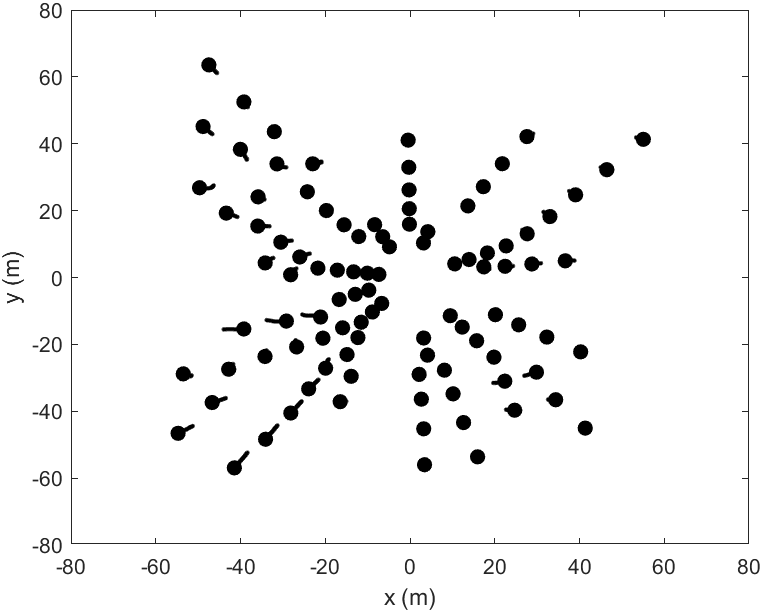}}
    \caption{One snapshot of radial pattern formation behavior among one hundred agents. As time progresses, the agents gradually form paths from random starting positions and stretch the paths until a stable equilibrium is reached.}
    \label{fig:path}
\end{figure}
In this section we have the robots forming a radial pattern. The pattern/path formation task is often interpreted in several different ways. In some cases, the agents are controlled to mimic the pheromonal stigmergy behavior as in ants and other insects~\cite{couzin2003ants}. Another version seen in robots with limited sensing ranges is the construction of “chains” where the “links” are communication connections between robots spread out in such a way that communication can be maintained throughout the chain and a destination can be reached that would be outside of any individual’s communication range~\cite{pereira2003communication}. Looking to mimic this behavior, the control structure and parameters were set to:
\begin{equation} 
\label{eqt:path_struct}
    \begin{bmatrix} 
        2.5 & -0.5 & 0 & -2.5\\ 
        0 & 0 & 5 & 0\\ 
        0.5 & 0 & 0 & -1 
    \end{bmatrix}
    \times 
    \begin{bmatrix}
        S_{N}\\ 
        S_{O}\\
        S_{d} . \theta\\
        1
    \end{bmatrix}
    \cdot 
    \begin{bmatrix} 
        \overrightarrow{V}_{N}\\
        \overrightarrow{V}_{ortho}\\
        \overrightarrow{V}_{O}
    \end{bmatrix}
    = 
    \overrightarrow{V_{c}}
\end{equation}
 A simulation with a group with 100 agents using such control law is shown in Fig.~\ref{fig:path}.  The scalars included in the structure for this application are the distance to the closest neighbor $S_{N}$, the distance to the origin of the map $S_{O}$, a constant bias, and a value found as the product of two more simple measurements $S_{P}$. The first value used in the product is the difference between the robot's distance to the origin and its closest neighbor's distance to the origin $S_{d}$. The second is the angular difference between the vector pointing from the origin to the agent and the vector pointing from the origin to the agent's closest neighbor $\theta$. This product was used to correct a sign mismatch when hand-tuning the parameters. The vector set for this application consisted of the unit vector pointing towards the agent's nearest neighbor $\overrightarrow{V}_{N}$, a unit vector orthogonal to the nearest neighbor vector $\overrightarrow{V}_{ortho}$, and the unit vector pointing towards the origin of the map $\overrightarrow{V}_{O}$.

\vspace{-.5mm}
\subsection{Collision Avoidance}
\begin{figure}[htbp]
   \centerline{\includegraphics[width=0.75\linewidth]{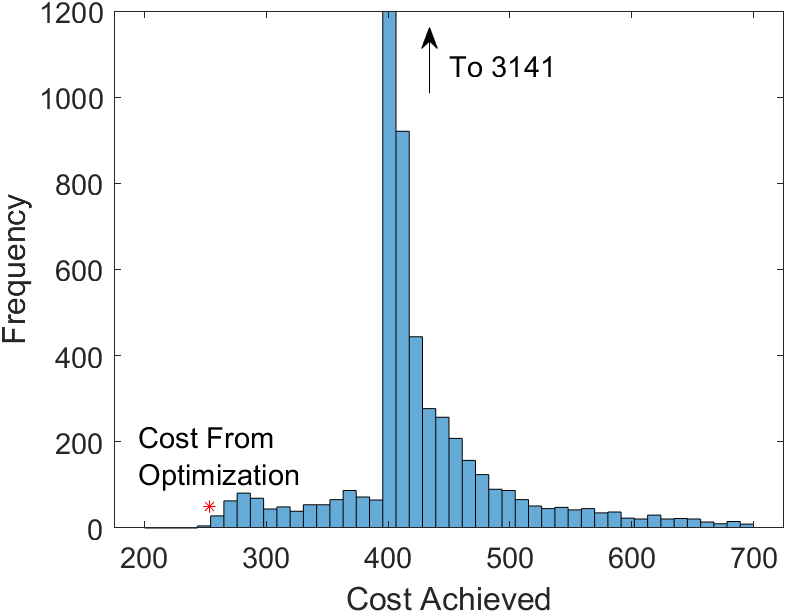}}
    \caption{Histogram showing results of a ten thousand iteration Monte-Carlo simulation with random parameter choice for the four agent antipodal swap controller. The star at the left extreme of the distribution indicates the cost achieved using optimization (253.4), which is lower than any cost achieved over ten thousand random iterations.}
    \label{fig:hist}
\end{figure} 

In contrast to the previous tasks where the parameters were hand-tuned, this subsection shows how the Bayesian optimization technique can be used to allow collision avoidance. 
A cost function was defined that penalized agents for getting close to one another and penalized agents each time step for not reaching their goal. 
The three scalars included in the controller structure were the straight line distance to the goal location, the inverse of the distance to the nearest neighbor, and a constant bias term. The four vectors included were the unit vector to the nearest neighbor, the unit vector to the goal, and orthogonal vectors to each. To test the method, a scenario was created with four agents performing an antipodal swap. 
In order to determine if the optimization function was able to find the global minimum of the given cost function, a Monte-Carlo simulation was run with ten thousand trials. For each trial, the parameters of the control structure were randomized and the simulation was run for one hundred time steps. The cost achieved by the optimization method was lower than the minimum achieved by any trial of the Monte-Carlo simulation as shown in Fig.~\ref{fig:hist}. The spike in the distribution at four hundred is an artifact of the four agents never reaching their goals when using the random parameters and being penalized one point each at every time step.

\subsection{Searching/Foraging}

\begin{figure}[tbp]
   \centerline{\includegraphics[width=0.77\linewidth]{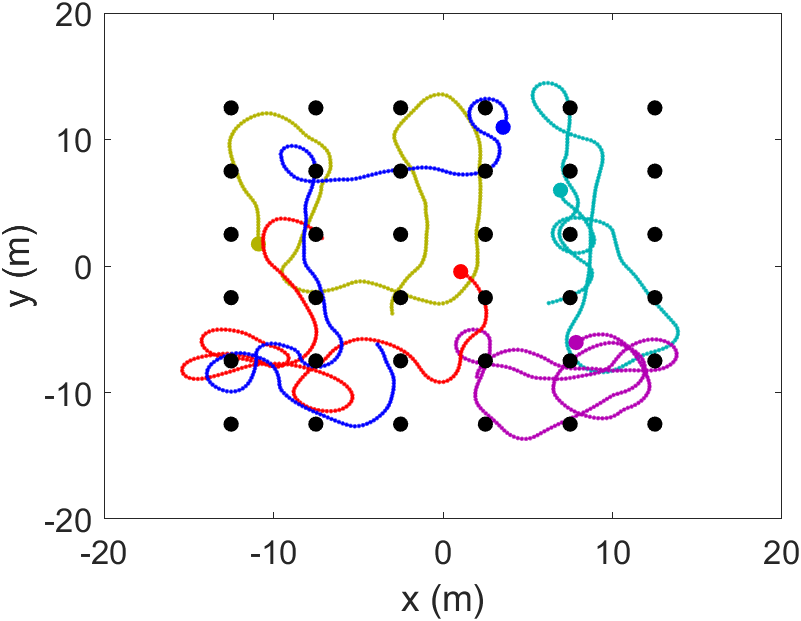}}
    \caption{Robots' paths during three hundred time steps of the searching/foraging task with search locations shown in black. The paths shown were generated using the optimized controller in contrast to the paths shown in Fig.~\ref{fig:hand_search}, obtained with a hand-tuned model.}
    \label{fig:search}
\end{figure}
In this section we solve the foraging/searching task using the complete methodology, including the Bayesian optimization step. For the problem formulation, a rectangular grid of 36 points was generated representing “search locations”. In order to formulate a representative cost function, each point on the search grid was given a counter that would increase each timestep until hitting a maximum. When an agent traveled close to a given point, its counter was reset to zero. At each timestep, the values of all the counters were added to the overall simulation cost. Four scalars and five vectors were included in the control structure. The scalars included the agent’s distance from the center of the map, the distance to the nearest neighbor, and the counter value of the closest search location, as well as a constant bias term. Like in previous examples, the distance to the map's center was divided by the map radius and raised to the fourth to decrease its importance near the center of the map. Also, the inverse of the distance to the nearest neighbor squared was found to increase repulsion utility. The vectors included in the control structure are a unit vector pointing towards the centroid of nearby search location weighted by their counters, the unit vector pointing towards the nearest neighbor, the unit vector pointing towards the nearest search location, the current velocity vector, and the unit vector pointing towards the origin of the map.

In order to get some intuition for the problem, a hand-tuned solution was created in addition to the solution produced using Bayesian optimization. The optimization technique was able to find a solution with similar efficiency to the hand-tuned solution. However, unlike in the collision avoidance simulation, the minimum achieved cost by the optimization technique was higher than that achieved using hand-tuning. Representative search paths with 5 agents are shown in Fig.~\ref{fig:hand_search} and Fig.~\ref{fig:search}.

\subsection{Real World Implementation}

\begin{figure}[ht]
\centering
   \includegraphics[width=0.95\linewidth]{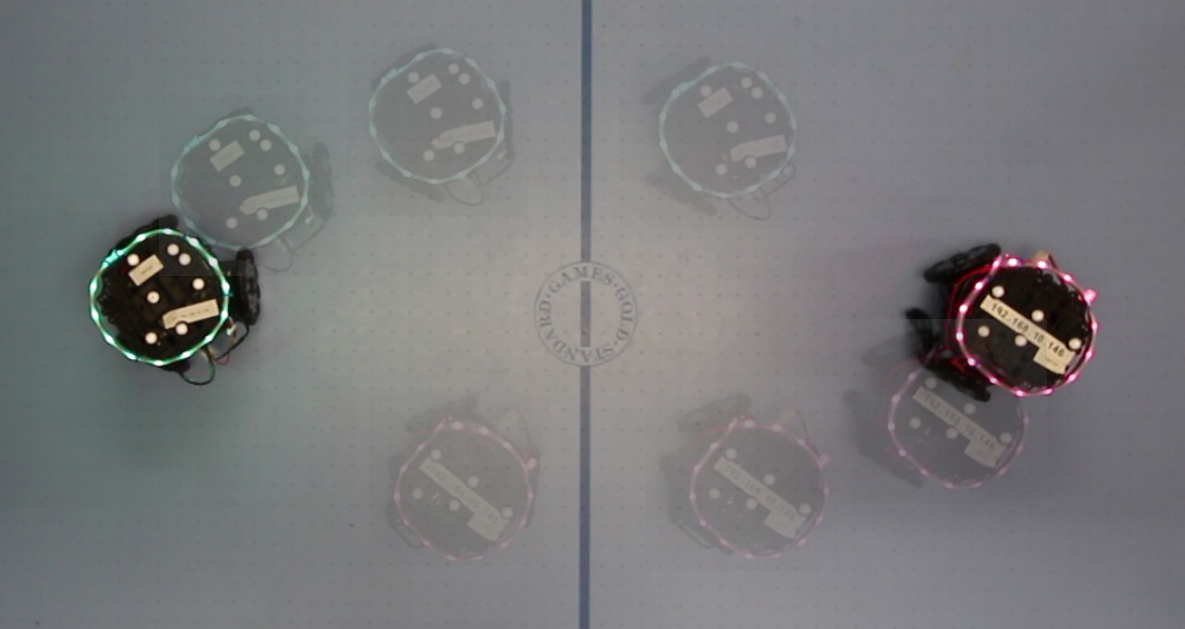}
    \caption{Several overlayed snapshots of two physical robots exhibiting an antipodal swap behavior using a hand-tuned version of the collision avoidance controller.}
    \label{fig:real_world}
\end{figure}

In addition to the simulation results, some experiments were performed on real-world holonomic robots. As shown in Fig.~\ref{fig:real_world}, the collision avoidance control structure was applied to two robots performing an antipodal swap. The control scheme was performed on a central computer with control velocities distributed and performed by the individual robots. Position information was collected using a Vicon motion capture system. In the cases tested, the robots were able to effectively avoid each other and reach their goal positions.

\vspace{-1mm}
\subsection{Discussion}
Five implementations of the proposed control framework have been described. In three cases, a hand-tuned parameter set was found that accomplished a multi-robot task that has been presented in prior literature, and, in two cases, parameter sets were found using Bayesian optimization. While the behaviors presented are not globally optimal as compared to behaviors possible using centralized approaches or non-linear controllers, the structures behind the different controllers are built upon a common framework that can be further expanded to suit other decentralized multi-robot tasks.

\section{Conclusion and Future Work}
This paper presented a general framework for decentralized multi-robot control. While the results presented in the paper may be preliminary, the five applications and the presented real-world experiment show the potential of the framework for future application and research. Given a meaningful set of measurements, a well-defined cost function for a simple task, and a tailored optimization method, the framework is able to find a set of parameters that will allow the group of robots to execute the desired task. A potential benefit of the framework, which will be explored in our future research, is the possibility of changing the team behavior in execution time by simply changing the set of parameters used.

There are still several challenges that need to be addressed to increase the utility of the framework and that are also part of our future work. In the current state, defining an effective cost function to use in the optimization is challenging and can lead to unwanted results. It is possible that techniques from reinforcement learning, such as reward shaping~\cite{laud2004theory} or inverse reinforcement learning~\cite{ng2000algorithms}, could be applied to solve this challenge. This would allow us, for example, to observe the behavior of a natural swarm and seek to find a cost function that will make the group mimic the natural behavior. Another challenge with this framework is choosing the appropriate set of measurements. Too many measurements increase the dimensional of search space that presents a barrier for any optimization technique, whereas too few measurements limit the complexity of the solution. We plan to solve this problem by applying known techniques for structure selection in system identification~\cite{aguirre1994some} or investigate compression-based approaches from the machine learning community that reduce the input size without loss of useful information~\cite{kingma2013auto}. In the future, we also plan to present theoretical proofs indicating that the use of Bayesian optimization indeed results in true global optimum with high probability. Furthermore, we expand the results by comparing the performance of Bayesian optimization against other black-box optimization techniques including particle-swarm optimization~\cite{kennedy1995particle}. 

Another correlated challenge that can benefit from similar techniques is deciding what form a measurement should take by choosing the best vectors and scalars to be considered. We noticed, for example, that some improvements have been seen by modifying scalar inputs, such as by taking their inverse, but these decisions have not been data-driven and would be better handled over to an expansion of the technique.


\section{Acknowledgments}

The authors would like to thank Dr. Jason Gross and Dr. Xi Yu for providing  guidance throughout the research process. We would like to also thank David Rubel and all of the other participating members of WVU's REU 2021 program for assisting with the formation of ideas and physical implementation of this research. 

\newpage
\bibliographystyle{IEEEtran}
\balance
\bibliography{bibliography.bib}

\end{document}